\documentclass[letterpaper, 10 pt, conference]{ieeeconf}
\IEEEoverridecommandlockouts
\title{Flip Stunts on Bicycle Robots using Iterative Motion Imitation}
\author{
    Jeonghwan Kim$^{1,2}$, 
    Shamel Fahmi$^{1}$, 
    Seungeun Rho$^{1,2}$, 
    Sehoon Ha$^{2}$, 
    and Gabriel Nelson$^{1}$%
    \thanks{
    $^{1}$RAI Institute, Cambridge, MA, USA. 
    $^{2}$Georgia Institute of Technology, Atlanta, GA, USA. 
    This work was done while J.~Kim and S.~Rho were at the RAI Institute.
    Project page: \href{https://imi-umv.github.io/}{\texttt{https://imi-umv.github.io/}}. 
    \tt{sfahmi@rai-inst.com, jkim3662@gatech.edu}.
    }
}

\usepackage[final]{hyperref} 
\hypersetup{colorlinks=true, citecolor=MidnightBlue, linkcolor=MidnightBlue, urlcolor=MidnightBlue}
\usepackage{cite} 
\usepackage{graphicx} 
\graphicspath{{./includes/}, {./includes/supplementary_includes/}}
\usepackage[]{units} 
\usepackage{xspace} 
\usepackage{amsfonts} 
\usepackage[dvipsnames]{xcolor} 
\usepackage{tabularx}
\usepackage{multicol}
\usepackage[nonumberlist, nogroupskip, section=section, numberedsection=autolabel]{glossaries} 
\newacronym{rl}{RL}{Reinforcement Learning}
\newacronym{imi}{IMI}{Iterative Motion Imitation}
\newacronym{umv}{UMV}{Ultra-Mobility Vehicle}
\newacronym{cmdp}{CMDP}{Constrained Markov Decision Process}
\newacronym{mdp}{MDP}{Markov Decision Process}
\newacronym{rsi}{RSI}{Reference State Initialization}
\newacronym{ppo}{PPO}{Proximal Policy Optimization}
\newacronym{mlp}{MLPs}{Multi-Layer Perceptrons}
\newacronym{rlhf}{RLHF}{Reinforcement Learning from Human Feedback}

\newcommand{\upper}{upper-body\xspace}
\newcommand{\lowerr}{lower-body\xspace}
\newcommand{\fork}{fork\xspace}
\newcommand{\frontwheel}{front-wheel\xspace}
\newcommand{\rearwheel}{rear-wheel\xspace}
\newcommand{\base}{bike-base\xspace}
\newcommand{\ig}{IsaacLab\xspace}

\newcommand{\ie}{i.e.,\xspace}
\newcommand{\fref}[1]{Fig.~\ref{#1}} 
\newcommand{\tref}[1]{Table~\ref{#1}} 
\newcommand{\boldSubSec}[1]{\noindent\textbf{#1.}}
\newcommand{\boldSubSecColon}[1]{\noindent\textbf{#1:}}
\newcommand{\bicycle}{bicycle\xspace}

\begin{document}
\maketitle

\begin{abstract}
This work demonstrates a front-flip on \bicycle robots via reinforcement learning, particularly by imitating reference motions that are infeasible and imperfect.  
To address this, we propose~\acrfull{imi}, a method that iteratively imitates trajectories generated by prior policy rollouts.
Starting from an initial reference that is kinematically or dynamically 
infeasible,~\acrshort{imi} helps train policies that lead to feasible and agile behaviors.
We demonstrate our method on~\acrfull{umv}, a \bicycle robot that is designed to enable agile behaviors.
From a self-colliding table-to-ground flip reference generated by a model-based controller,
we are able to train policies that enable ground-to-ground and ground-to-table front-flips.
We show that compared to a single-shot motion imitation,~\acrshort{imi}
results in policies with higher success rates and can transfer robustly to the real world. 
To our knowledge, this is the first unassisted acrobatic flip behavior on such a platform.

\end{abstract}


\section{INTRODUCTION}
Wheeled and legged robots have recently demonstrated remarkable dynamic capabilities, driven in large part by advances in~\gls{rl} and motion imitation~\cite{hwangbo2019learning,lee2024learning, vollenweider2022advanced, cheng2024extreme, zhuang2023robot, hoeller2024anymal, miller2023reinforcement}. 
By tracking demonstrations from 
motion capture~\cite{peng2018deepmimic, peng2021amp}, 
animal locomotion~\cite{peng2020learning, yoon2025spatio}, 
or model-based controllers~\cite{miller2023reinforcement, youm2023imitating, kang2023rl+, jenelten2024dtc}, 
robots can learn parkour and agile behaviors. 
However, if the original motion references are dynamically or kinematically infeasible,
the imitation policy may fail to train or lead to unsafe behaviors not suitable for real-world deployment~\cite{yoon2025spatio}.
Blindly tracking these infeasible references may exceed the robot's torque or joint limits or cause self-collisions.


Prior works have explored 
learning from imperfect demonstrations by reweighting trajectories according to confidence in their optimality~\cite{wu2019imitation}, 
filtering out low-quality segments to extrapolate higher-quality behaviors~\cite{brown2019extrapolating}, 
or using adversarial techniques for partial demonstrations~\cite{li2023learning}. 
While effective to some extent, these approaches still treat imperfections as liabilities to be corrected offline and as a result, 
the final policy is still bounded by the static quality of the dataset. 

In this work, we introduce~\gls{imi}, 
an extension of motion imitation that iteratively transforms infeasible trajectories into feasible and agile behaviors. 
\gls{imi} begins by imitating an initial reference, which may be generated through manual control without safety constraints, trajectory optimization, or hand-crafted drawing, 
and is often physically infeasible. 
Once a policy is trained to follow this reference, 
its rollout is treated as a new reference for the next stage of training. 
Through this recursive process, imperfections in the reference are progressively removed, 
which allows the next policies to better satisfy safety-critical constraints. 
This allows us to 
use simple reward functions and constraints without relying on extensive reward shaping,
and it enables feasible and agile behaviors.
%

\begin{figure}[t!]\centering
\includegraphics[width=0.86\columnwidth]{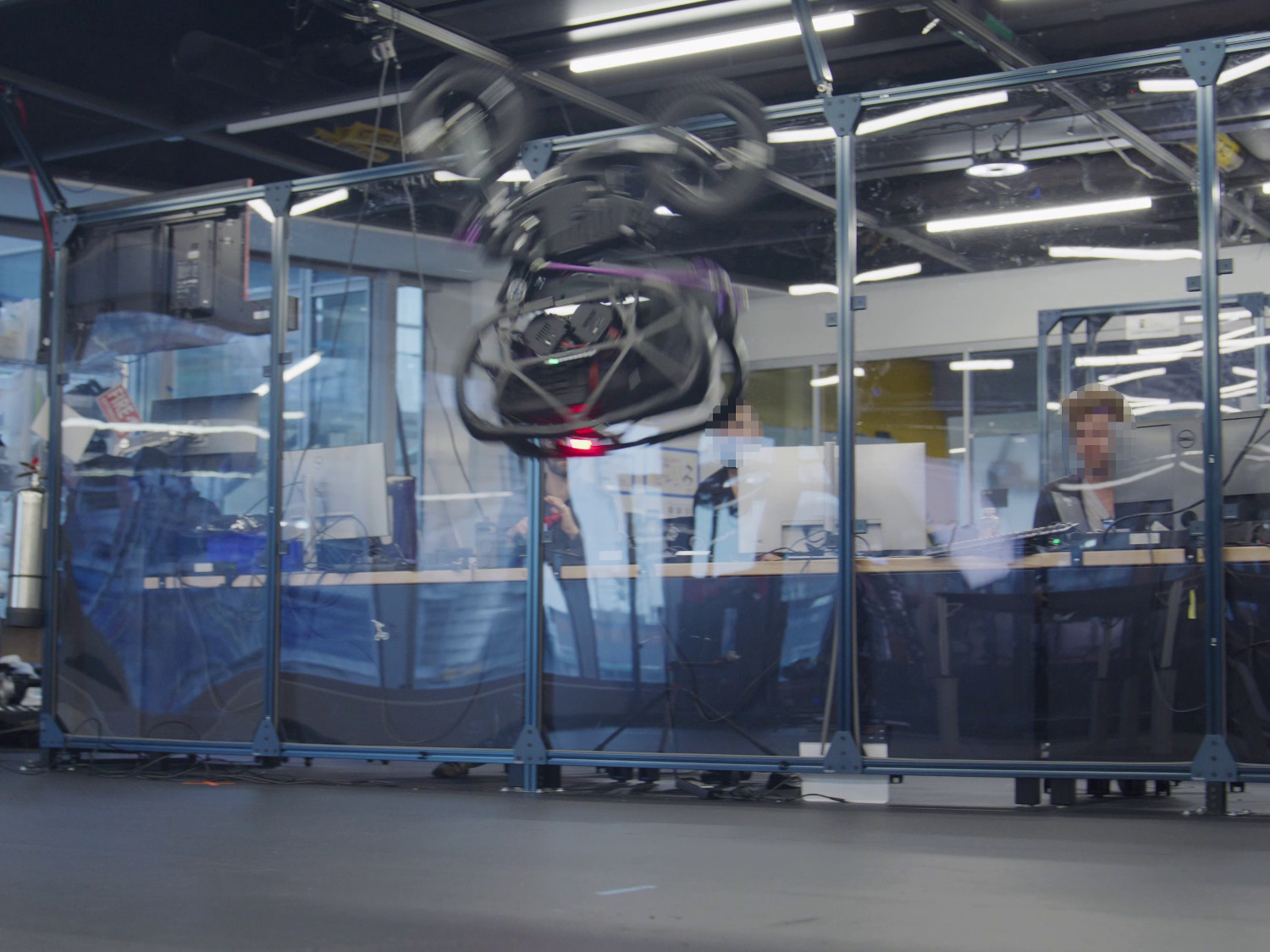}		
\caption{\acrshort{umv} executing a front-flip learned via~\acrlong{imi}.}
\label{fig_res4}
\end{figure}

We deploy \gls{imi} on \gls{umv}~\cite{RAIInstitute2026UMV}, a \bicycle robot equipped with a large articulated mass for acrobatic maneuvers as shown in \fref{fig_res4}.
We focus on a front-flip, as its agility allows us to push the system to its mechanical limits.
Beginning with an initial flip-down trajectory with simple imitation rewards, this iterative process allows the policy to generalize to scenarios beyond what the original reference was designed for.
We demonstrate this extended agility with ground-to-ground flips and ground-to-table flip-ups.
We further confirm the robustness of the learned behaviors through hardware deployment.

To sum up, our contributions are:
\begin{itemize}
    \item \acrfull{imi}, a method that refines imperfect references into agile policies via iterative imitation, yielding highly agile behaviors from simple tracking rewards.
    \item Analysis of performance improvement and trajectory for both original flip stunts and adapted flip-up scenarios.
    \item Hardware demonstration of acrobatic flip stunts on the \gls{umv} robot, validating the approach on a platform with unique dynamic stability challenges.
\end{itemize}


\section{RELATED WORK}
\subsection{Reinforcement Learning for Agile Locomotion}
\gls{rl} has emerged as a dominant paradigm for synthesizing policies that achieve dynamic locomotion skills beyond the reach of classical control. Recent works have enabled quadrupeds to reach record speeds~\cite{li2025reinforcement, margolis2024rapid, miller2025high}, execute agile parkour~\cite{cheng2024extreme, kim2025high, rho2025unsupervised}, and generalize across terrains~\cite{lee2020learning, he2025attention}. These approaches are often ``reference-free,'' relying solely on carefully engineered rewards. While powerful, the reliance on hand-tuned multi-term rewards can be a bottleneck, limiting scalability to diverse and extreme behaviors.

Motion imitation methods alleviate this challenge by leveraging demonstrations such as motion capture~\cite{peng2018deepmimic, peng2020learning}, video~\cite{peng2018sfv}, or trajectory optimization~\cite{fuchioka2022opt, liu2024opt2skill}. 
Along with publicly available motion capture datasets~\cite{mahmood2019amass,harvey2020robust}, motion imitation methods have accelerated the research in humanoid whole-body control~\cite{he2025asap,chen2025gmt,he2024omnih2o}. However, such datasets not only have motion artifacts~\cite{Luo2023PerpetualHC}, but the process of retargeting to robot morphologies also leads to artifacts, resulting in physically infeasible references~\cite{ze2025twist,xie2025kungfubot}.

\subsection{Learning from Imperfect References}
In practice, reference trajectories are rarely ideal, often suffering from noise, morphological mismatches, or dynamic infeasibility~\cite{yoon2025spatio}. 
Existing methods address this by treating imperfect demonstrations as a static dataset. 
For instance, 
trajectories may be weighted or ranked based on scores or demonstration quality~\cite{wu2019imitation,brown2019extrapolating,cao2021learning}. 
Trajectory optimization can also be used to refine the reference offline~\cite{yoon2025spatio}. 
While useful, these techniques cannot refine demonstrations through interaction, 
and optimization-based methods often require an accurate dynamics model. 
Moreover, these refinements cannot generate references that exceed the agility of the original demonstrations~\cite{ranawaka2025sail}. 

An alternative is the adversarial imitation framework~\cite{peng2021amp}, which can produce natural behaviors on real robots~\cite{peng2020learning, wu2023learning}, even from partial demonstrations~\cite{li2023learning}. 
Yet, these methods are known to be unstable during training and highly sensitive to hyperparameter tuning~\cite{peng2018variational, li2025featurebased}.

Our work departs from this perspective: rather than coping with reference imperfections, \gls{imi} recursively transforms them into feasible and robust guides for policy learning. 
This is done in the same framework, without re-tuning, and without designing complicated rewards. 


\subsection{Learning Bicycle Stunts}
Bicycle control
has long served as a benchmark for both model-based and learning-based robotics. 
Linear~\cite{xiong2024steering} and nonlinear~\cite{cui2020nonlinear} 
control strategies were studied on basic tasks such as balancing and target reaching. 
Early works on~\gls{rl} for bicycle control demonstrated that careful reward shaping can solve basic driving tasks~\cite{randlov1998learning}.

Subsequent research has leveraged~\gls{rl} to push bicycle capabilities toward 
driving in complex environments. 
Hierarchical~\gls{rl} has been used 
to achieve robust path tracking and balancing on rough, unstructured terrain~\cite{zhu2023deep}. 
Others have integrated their~\gls{rl} frameworks 
combining path planning, trajectory tracking, and balancing to navigate narrow corridors~\cite{zheng2023reinforcement}. 
This trend extends beyond bicycles, as demonstrated by Baltes et al.~\cite{baltes2023deep}, who trained a humanoid robot to steer and balance on a commercial scooter.

Beyond driving, dynamic stunts such as jumping have been studied. 
Learning-based methods have been successfully applied to control ramp jumps by optimizing for flight attitude and landing~\cite{zheng2022ramp}, while model-based approaches have provided complementary insights into the underlying physics using techniques like inverse kinematics and Bayesian optimization~\cite{wang2024bayesian}.
Tan et al.~\cite{tan2014stunts} significantly raised the bar of bicycle agility, 
employing neuro-evolution to successfully learn a diverse repertoire of acrobatic stunts, 
including bunny hops, wheelies, endos, and pivots.

While these studies showcase a range of dynamic behaviors, most of their results are confined to simulation, lacking validation on physical hardware. Our work builds on this foundation by not only tackling acrobatic flips—a maneuver of greater dynamic complexity than previously demonstrated—but also by successfully deploying the learned policy on a real-world robotic platform.


\begin{figure}\centering
\includegraphics[width=0.88\columnwidth]{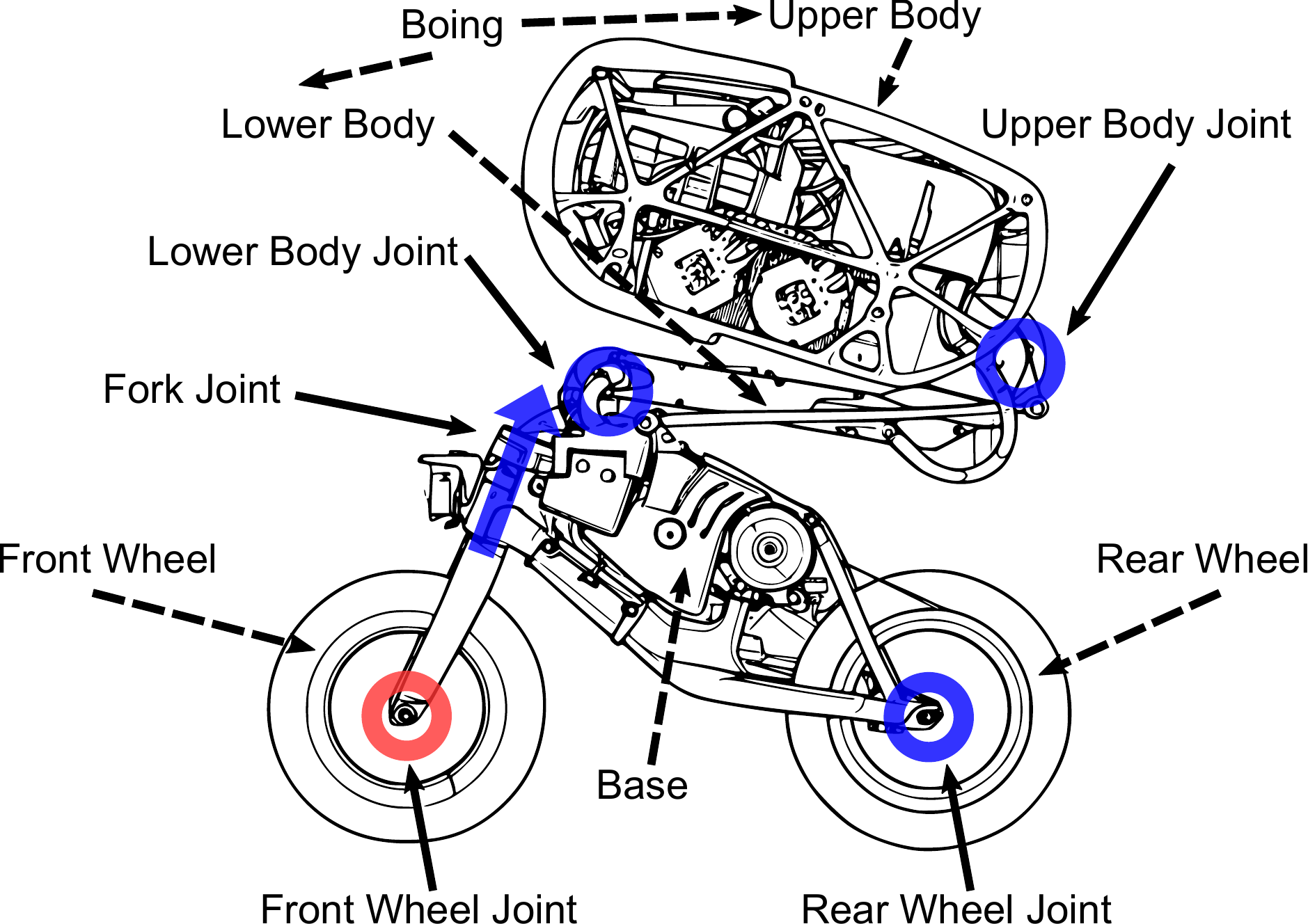}
\caption{A 2D overview of the \gls{umv} model used in this work.}
\label{fig:UMVMini}
\end{figure}

\section{The~\acrfull{umv} Robot} 
\gls{umv} is a custom-built, two-wheeled robot with a bike base comparable in size to a children's bicycle (Fig. \ref{fig:UMVMini}). A key feature that distinguishes the \gls{umv} from a conventional bicycle is a significant articulated mass, referred to as~\emph{boing}, mounted atop the \base frame.

The model of \gls{umv} used in this work consists of five joints and six links. 
The links are \upper, \lowerr, \base, \fork, \frontwheel, and \rearwheel. 
The joints are \upper, \lowerr, \fork, \frontwheel, and \rearwheel.
\gls{umv} has four actuated joints in total. 
The \upper and \lowerr joints control the pitch of the upper body, and are the main joints to inject momentum to perform parkour behaviors. 
The fork joint controls the steering angle (yaw) of the front wheel. 
The rear-wheel provides driving torque.
Consistent with standard bicycle design, the front wheel remains passive and unactuated.

Boing (the \upper and \lowerr links) houses the batteries and joint motors and is connected to the \base by two parallel revolute joints (the \upper and \lowerr joints). 
The axes of these joints are parallel to the wheel axles, allowing the \upper to pitch forward and backward relative to the \base. 
This mechanism enables the robot to dramatically shift its center of mass, a critical capability for generating the angular momentum required for acrobatic maneuvers. 


\section{METHOD}

\subsection{The \acrfull{imi}}
\begin{figure}[t!]
    \centering
    \includegraphics[width=0.95\columnwidth]{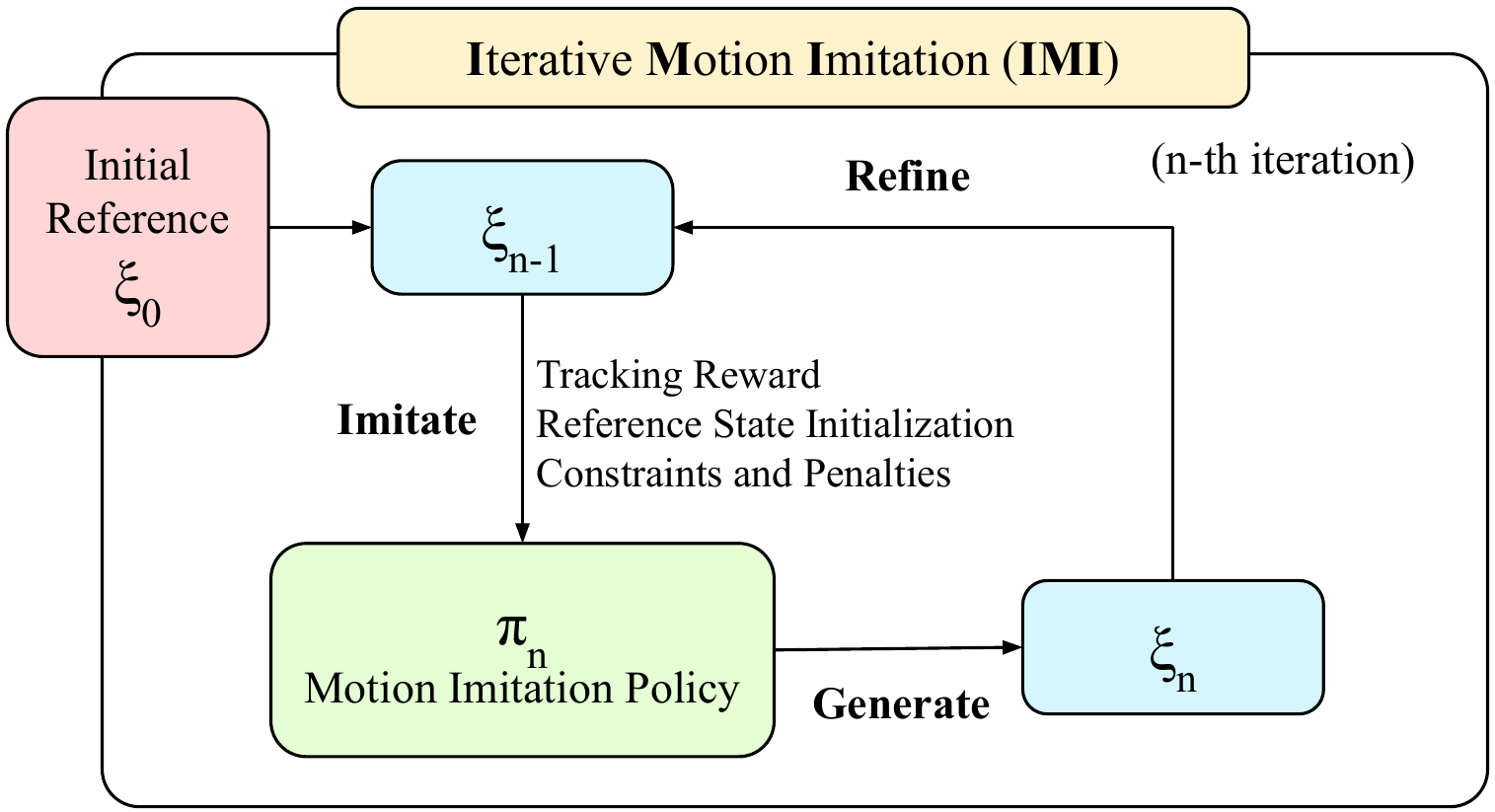}
    \caption{An overview of~\gls{imi}. 
    The first iteration starts with an initial reference~$\xi_0$ used to train policy~$\pi_1$.
    Then, reference~$\xi_{n-1}$ generated by policy~$\pi_{n-1}$ 
    becomes the new more feasible reference used to train a new policy~$\pi_n$. 
    This loop continues, progressively refining the motion until a high-performance, hardware-deployable policy is achieved.}
    \label{fig:imi_framework}
\end{figure}

The goal of~\gls{imi}, depicted in~\fref{fig:imi_framework},  is to acquire robust and agile skills from an infeasible trajectory. 
The framework iteratively refines an initial, imperfect reference until a policy exhibiting the desired feasible behavior is achieved. This process involves three key steps
\begin{enumerate}
\item \textbf{Imitate:} The policy is trained using constrained \gls{rl} to track the current reference trajectory.
\item \textbf{Generate:} The learned policy is executed in simulation to generate a new, physically plausible trajectory.
\item \textbf{Refine:} The generated trajectory is set as the new reference for the next imitation cycle. 
\end{enumerate}


The kinematic reference trajectories are used exclusively during training and are progressively improved with each cycle. 
Experts can also manually trim the trajectory during the refine stage when the desired motion differs from the reference motion, such as adapting a flip-down trajectory to a flat-to-flat flip.

The learned policy is purely reactive, taking only proprioceptive state observations and a phase variable as input to produce joint-level PD targets, which are then tracked by a high-frequency low-level controller.

\gls{imi} enhances the two primary guidance mechanisms of motion imitation, dense tracking rewards and \gls{rsi}~\cite{peng2018deepmimic}, by progressively improving the reference trajectory itself. 
This iterative refinement makes the tracking reward more informative. Initially, an infeasible reference creates a conflicting objective, forcing the policy to deviate from the reference to satisfy physical constraints. 
As \gls{imi} generates more plausible references, the policy can achieve high tracking fidelity while respecting these constraints. Concurrently, \gls{rsi} becomes more effective. Initializing from a refined, physically achievable reference places the agent in meaningful states that are closer to high-reward regions, accelerating learning and exploration.


\subsection{Initial Reference Trajectory Generation}
The initial flip reference can be generated in various ways, but we opt to use a hand-crafted model-based controller without safety constraints in simulation. 
The robot starts on top of a table and moves forward using a driving controller to build forward linear momentum. 
As the robot reaches the edge of the table, a whole-body controller is used to track a certain angular momentum that is tuned to successfully flip down the table. 

This orchestrated scenario was chosen intentionally. 
Starting on top of a table, reaching a certain forward momentum, and tracking and tuning a whole-body controller 
were engineered to provide an extended flight phase, giving the robot ample time and momentum to complete a 360-degree rotation.

While this procedure yields a reference trajectory that achieves a flip, it is not executable without switching mid-air to a separate landing controller. 
Moreover, this does not meet our desired agility of flipping from flat ground and also violates the desired safety limits. 
Thus, the initial reference serves only as a rough demonstration, and our method aims to refine this imperfect reference into an end-to-end deployable policy.




\subsection{Problem Formulation}
We formulate the motion imitation task as a \gls{cmdp}, defined by the tuple $(\mathcal{S}, \mathcal{A}, \mathcal{P}, r, \mathcal{C}, \gamma)$. Here, $\mathcal{S}$ is the state space, $\mathcal{A}$ is the action space, $\mathcal{P}(s_{t+1}|s_t, a_t)$ is the state transition probability, $r(s_t, a_t)$ is the reward function, $\mathcal{C} = \{c_1, ..., c_k\}$ is a set of $k$ constraint functions, and $\gamma \in [0, 1)$ is the discount factor.
The objective is to find an optimal policy $\pi^*$ that maximizes the expected discounted sum of future rewards while satisfying a set of constraints. These constraints, $c_i(s_t, a_t)~\leq~0$, represent the physical and operational limits of the robot. 
The full objective is:
\begin{equation}
\pi^* = \arg\max_{\pi} \mathbb{E}_{\xi \sim \pi} \left[ \sum_{t=0}^{T} \gamma^t r(s_t, a_t) \right]
\end{equation}
\begin{equation}
\text{subject to } \quad c_i(s_t, a_t) \leq 0, \quad \forall i \in \{1,...,k\}, \forall t
\end{equation}
We handle the constraints by terminating the episode upon any violation as used in~\cite{chane2024cat}. This strategy effectively transforms the~\gls{cmdp} into a standard~\acrshort{mdp}, which we solve with RL: with a positive per-step reward design, any policy that violates constraints will inherently achieve a lower expected cumulative reward.
Each iteration in \gls{imi} is solved using \gls{ppo}~\cite{schulman2017proximal}. 

\subsection{Control, Observations, and Actions}
We use \ig as our training environment~\cite{mittal2023orbit}. 
The policy operates at~\unit[50]{Hz}, outputting joint PD targets. 
These targets are tracked by a low-level PD controller running at~\unit[200]{Hz} in simulation, and at~\unit[1]{kHz} on the real robot.

The proprioceptive observations $o_t \in \mathcal{S}$ are defined as
\begin{equation}
o_t = [q, \dot{q}, \omega, g, \theta, a_{t-1}] \in \mathbb{R}^{18}
\end{equation}
where $q \in \mathbb{R}^3$ are the actuated joint positions, 
$\dot{q} \in \mathbb{R}^4$ are the actuated joint velocities, 
$\omega \in \mathbb{R}^3$ is the base angular velocity, 
$g \in \mathbb{R}^3$ is the projected gravity vector in the robot's base frame, 
and $a_{t-1}\in \mathbb{R}^4$ is the previous action. 
The observations exclude the rear wheel's position, as it is an unbounded, continuously rotating joint. The phase variable $\theta \in \mathbb{R}^1$ increases linearly from 0 to 1 over the duration of the reference trajectory. The total episode duration is longer than the reference trajectory, and during this extra time, $\theta$ is held at 1, requiring the policy to learn a stable balancing behavior after the touchdown.


The policy outputs a four-dimensional action vector~$a_t~\in~\mathbb{R}^4$, corresponding to each joint's PD targets:
\begin{equation}
a_t = [a_{\text{\upper}}, a_{\text{\lowerr}}, a_{\text{\fork}}, a_{\text{\rearwheel}}]
\end{equation}
Each action is scaled by its corresponding action scale. 
The action of the \rearwheel joint is defined as a velocity setpoint, 
while the actions of the \upper, \lowerr, and \fork joints are defined as position setpoints. 
The setpoints are tracked by a PD controller with desired joint torques $\tau$
\begin{equation}
\tau = k_p(q_{\text{des}}- q) + k_d(\dot{q}_{\text{des}} - \dot{q})
\end{equation}
where~$q_{\text{des}}$ and ~$\dot{q}_{\text{des}}$ are the desired position and velocity setpoints for every joint.

{
\setlength{\tabcolsep}{3pt} 
\begin{table}[t]
\caption{Reward Structure with Tolerance}
\label{tab:reward_structure}
\centering
\resizebox{\columnwidth}{!}{%
\begin{tabular}{lccc}
\hline
\textbf{Component} & \textbf{Formulation} & \textbf{Weight} & \textbf{Tolerance} \\
\hline
\multicolumn{4}{c}{\textit{Tracking Rewards}} \\
Base Position & $\exp(-\alpha_{\text{base}} \| p_{\text{base}} - p_{\text{base}}^{\text{ref}} \|^2)$ & 4.0 & 0.4 \\
Base Orientation & $\exp(-\alpha_{\text{ang}} \| d_\text{angle}(q_{\text{base}},q_{\text{base}}^{\text{ref}}) \|^2)$ & 20.0 & 0.8 \\
Joint Positions & $\exp(-\alpha_{\text{joint}} \| q_{\text{joint}} - q_{\text{joint}}^{\text{ref}} \|^2)$ & 1.0 & 0.1 \\
[2pt]\hline
\multicolumn{4}{c}{\textit{Penalty}} \\
Action Smoothness & $\| a_t - a_{t-1} \|^2$ & -1e-5 to -1e-3 & --- \\
Fork Velocity & $\dot{q}_{\text{fork}}^2$ & -0.001 & --- \\
Contact Force & $\| F_{\text{contact}} \|^2$ & -1e-6 & 350 N \\
Body Joint Limits & $\mathbb{I}(q_{\text{joint}} \notin [q_{\text{min}}, q_{\text{max}}])$ & -1.0 & --- \\
[2pt]\hline
\multicolumn{4}{c}{\textit{Post Tracking Terms}} \\
Default Joint Positions & $\exp(-\alpha_{\text{pd}} \| q_{\text{body joint}} \|^2)$ & 1.0 & --- \\
Jitter Penalty & $\sum \dot{q}_{\text{joint}}^2 + \sum |\dot{q}_{\text{joint}}|$ & -1e-4 & --- \\ 
Velocity Penalty & $\| v_{\text{base}} \|^2$ & -3.0 & 0.5 m/s \\
[2pt]\hline
\end{tabular}
}%
\end{table}
}
\subsection{Rewards}
The reward is a weighted sum of tracking and penalty terms, detailed in~\tref{tab:reward_structure}. 
The tracking rewards encourage the policy to follow the reference motion, 
and penalties discourage behaviors that are unsuitable for hardware execution. 
We do not track the \fork and \rearwheel joint positions to grant the policy more freedom to discover strategies for balancing and momentum generation.
We clip the tracking errors to certain tolerances so that the maximum reward can be achieved without the need to track the exact reference.
After successfully tracking the reference (i.e., when the phase variable $\theta=1$), the \emph{post-tracking terms} are triggered to promote a still, balanced state by rewarding a default joint posture while penalizing base velocity and joint jitter.

\subsection{Constraints and Terminations}
To ensure the safe deployment of learned policies on hardware, we enforce critical physical limits as constraints. Similar to~\cite{chane2024cat}, we implement these constraints through termination conditions. Unlike approaches that use soft terminations to allow a policy to learn recovery behaviors, we employ hard terminations. This is because a constraint violation in our task represents an irrecoverable failure state. For instance, exceeding the current limit triggers a protective hardware shutdown, while an excessive touchdown velocity is an instantaneous event that can cause catastrophic hardware fracture.
Based on that, an episode is terminated if any of the following conditions are met:
\\
\boldSubSecColon{Touchdown Velocity} The landing vertical velocity of the wheels exceeds a certain threshold.\\
\boldSubSecColon{Mechanical Power} The total mechanical power of the joints exceeds a certain threshold~$\sum \tau \dot{q}$. This serves as a proxy for the peak current limit.\\
\boldSubSecColon{Joint Position} The bounds of the joint position limits are reached. This prevents self-collisions and singularities.\\
\boldSubSecColon{Motor Torques} The motor torques exceed their torque limit.\\
\boldSubSecColon{Wheel Velocity} The wheel velocity exceeds a safety limit.\\
\boldSubSecColon{Early Termination} The robot's \base position or orientation deviates too much from the reference trajectory.
\boldSubSecColon{Ground Collision} The \upper, \lowerr, or \base bodies collide with the ground. 

To encourage early exploration during training, we introduce these hard constraints via a curriculum. The training process begins with relaxed constraint boundaries that are gradually tightened as the policy improves. This approach allows the agent to first learn the fundamental task before refining its behavior to operate within the strict physical limits, thereby preventing the learning process from stagnating. 

Constraints can also be gradually introduced during different~\gls{imi} iterations. For instance, we only enforce the joint position limits at the first iteration, leaving other constraints to be tightened in future iterations. For further iterations, the joint position limits were enforced from the beginning of training, and the rest of the terminations are applied as a curriculum.

\subsection{Sim2Real and Policy Training}
\boldSubSec{\acrlong{rsi}} We use~\gls{rsi} to expose the policy to critical states early in training~\cite{peng2018deepmimic}. 
50\%~of the episodes begin from an initial state and the remaining 50\% are initialized at random states sampled from the reference trajectory.
We also ignore the last~10\% of the data during touchdown.

\begin{figure*}[t!]
    \centering
    \includegraphics[width=0.9\textwidth]{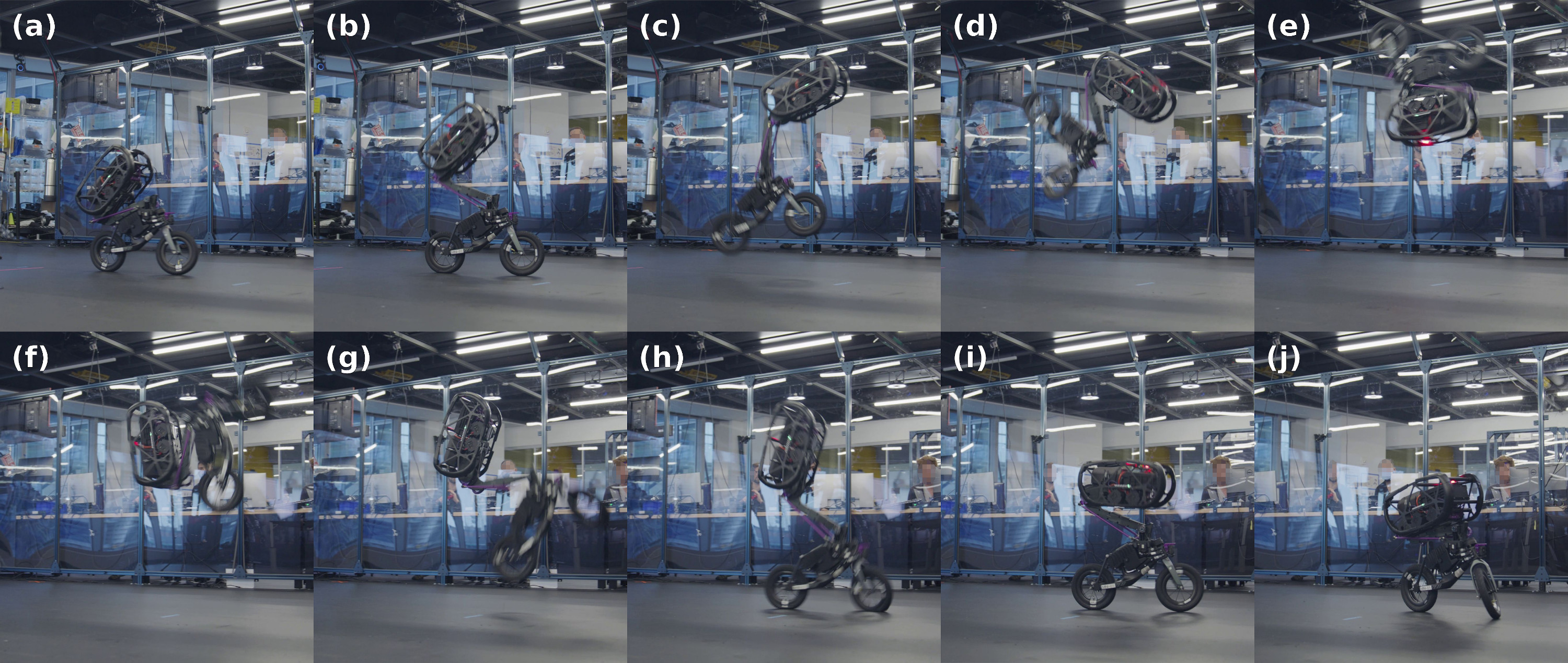}
    \caption{Overlayed screenshots of \gls{umv} performing a front flip.}
    \label{fig:flip_overlay}
\vspace{-5mm} 
\end{figure*}

\boldSubSec{Domain Randomization} 
To bridge the sim-to-real gap, we apply domain randomization to various physical and system parameters during all stages of \gls{imi}. 
Specifically, we randomize physical properties such as mass, friction, and motor strength, as well as control-related parameters including actuator gains and actuation delay. 
We also introduce observation noise to joint positions, velocities, angular velocity, and projected gravity, and further apply external disturbances to the body velocity.

Furthermore, to handle varied contact scenarios such as premature ground impact, we add terrain randomization. In 50\% of training episodes, we introduce a step obstacle with a height uniformly sampled from the range [\unit[0]{m}, \unit[0.2]{m}]. Collectively, these randomizations encourage the policy to learn behaviors that are robust to modeling errors and unmodeled dynamics.

\boldSubSec{Network Architecture}
The actor and critic networks are implemented as \gls{mlp} with ELU activation functions
and three hidden layers of size~[512, 256, 128] and~[512, 512, 256], respectively.
In addition to the actor's observations, the critic observes the \base's linear velocity and global position as privileged information during training.
The policies are trained for 15,000 iterations. 

\section{Results}

\begin{figure}[t]
    \centering
    \includegraphics[width=1.0\columnwidth]{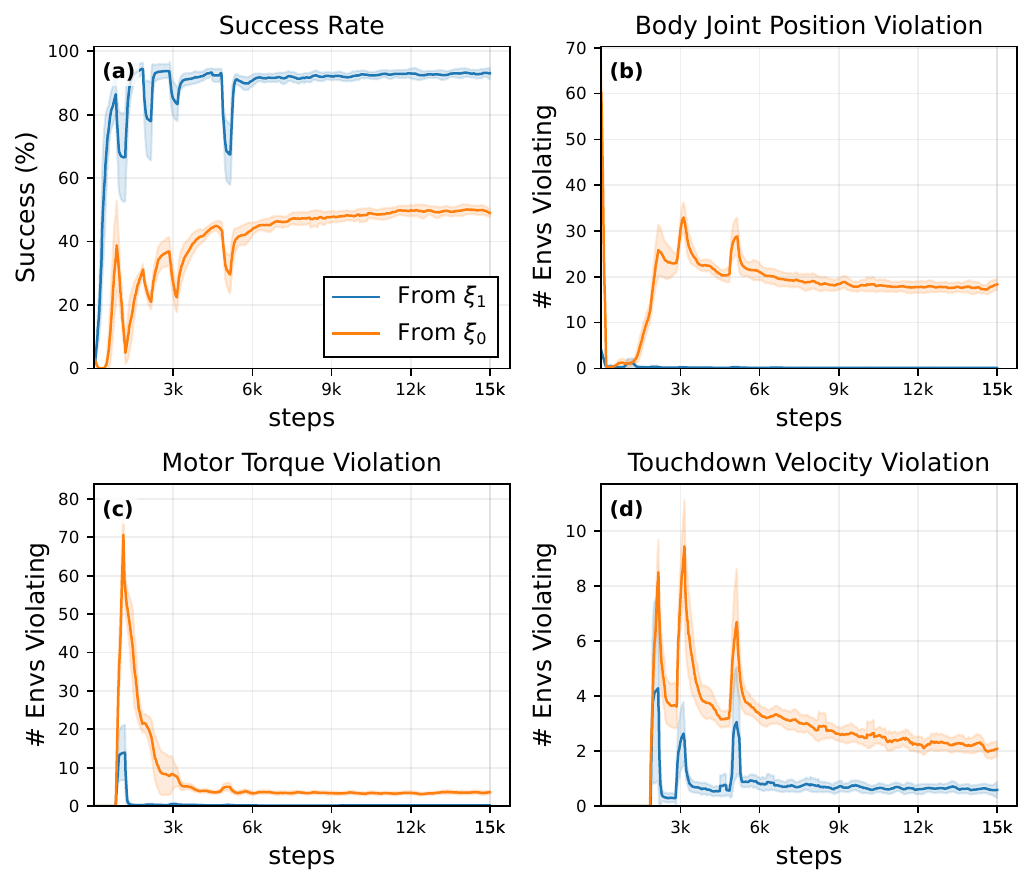}
    \caption{Comparison of training from scratch~($\xi_0$) versus using refined reference~($\xi_1$). 
    Each run uses 3 seeds. The solid line is the mean and the shaded area is the standard deviation.}
    \label{fig:ablation_wandb}
\end{figure}

We evaluate~\gls{imi} through simulations and experiments on~\gls{umv}, and show that it can transform an initially infeasible reference trajectory into a robust policy that transfers to the real world as shown in \fref{fig:flip_overlay}. 
We first detail the importance of iterative imitation on the sim-to-real transfer of our front-flip policy. 
Then, we conduct an ablation study to analyze the effect of iterative refinement, showing its ability to adapt to a harder maneuver such as a front-flip onto a table.

\subsection{Refinement of Flip Trajectories}\label{results_sec_1}
\begin{figure}[t]
    \centering
    \includegraphics[width=1.0\columnwidth]{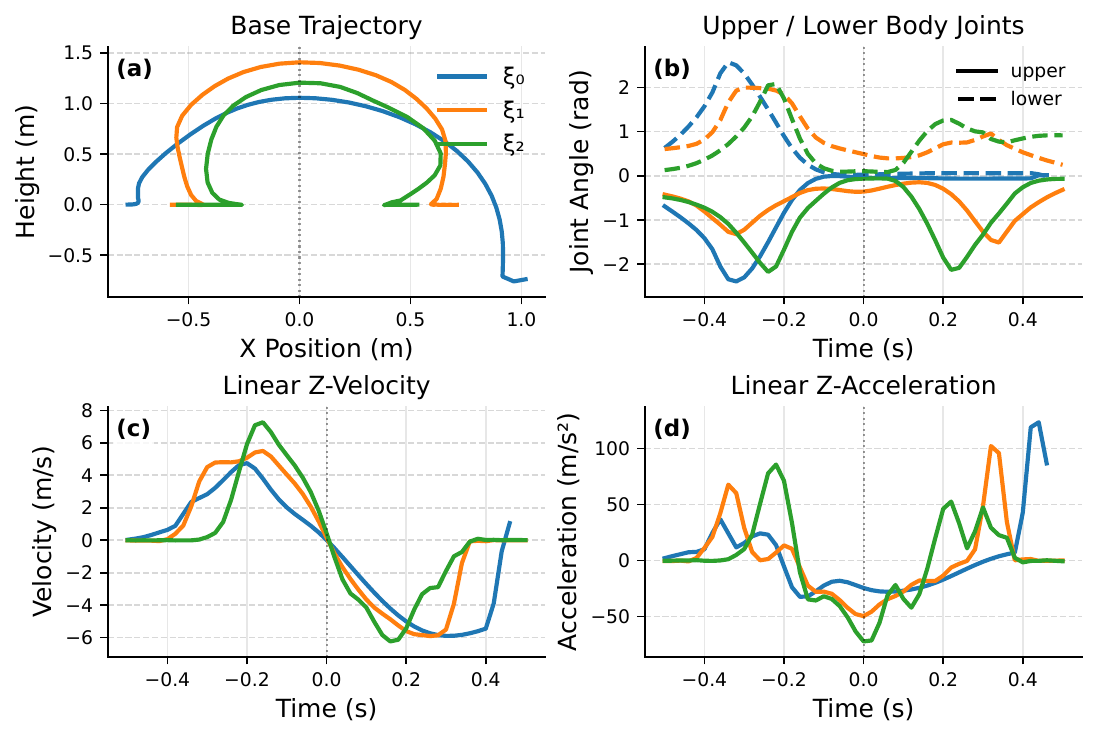}
    \caption{%
    The evolution of~trajectories~$\xi$ over two iterations.
    The time is aligned at the peak height~($t=0$) and the interval is~$\pm 0.5$~s around that peak.
    (a) \base X-Z position, 
    (b) \lowerr and \upper joint angles,
    (c) \base vertical velocity, and
    (d) \base vertical acceleration. 
    }
    \label{fig:trajectory_refinement}
\end{figure}

We train~\gls{imi} with two iterations. 
The first iteration trains a policy~$\pi_1$ with the initial trajectory generated by the model-based controller~(\ie $\xi_0$).
The second iteration trains a policy~$\pi_2$ with the trajectory generated from the first iteration~(\ie $\xi_1$).
Using the policy from the second iteration~$\pi_2$, we generate a third trajectory for comparison~(\ie $\xi_2$).

Since the initial reference~$\xi_0$ starts with the robot on a table, 
we translated the base pose so that the robot is initialized on the ground. For all references in the iteration, we trim the trajectory from the moment the robot acquires forward velocity until ground touchdown.

~\fref{fig:trajectory_refinement} illustrates the evolution of these trajectories~$\xi$. 
The initial trajectory~$\xi_0$ exhibits undesired behavior after the peak height: as shown in ~\fref{fig:trajectory_refinement}(b), the robot reaches its joint limits, leading to a self-collision. The landing is also unregulated, with the horizontal impact velocity and acceleration (~\fref{fig:trajectory_refinement}(c,d)) being the largest among the three iterations.  

In contrast, $\xi_2$ demonstrates improved landing characteristics over $\xi_1$. Its peak horizontal velocity and acceleration at touchdown are reduced (\fref{fig:trajectory_refinement}(c,d)), leading to a smoother and less damaging impact.


\begin{figure}[t]
    \centering
    \includegraphics[width=1.0\columnwidth]{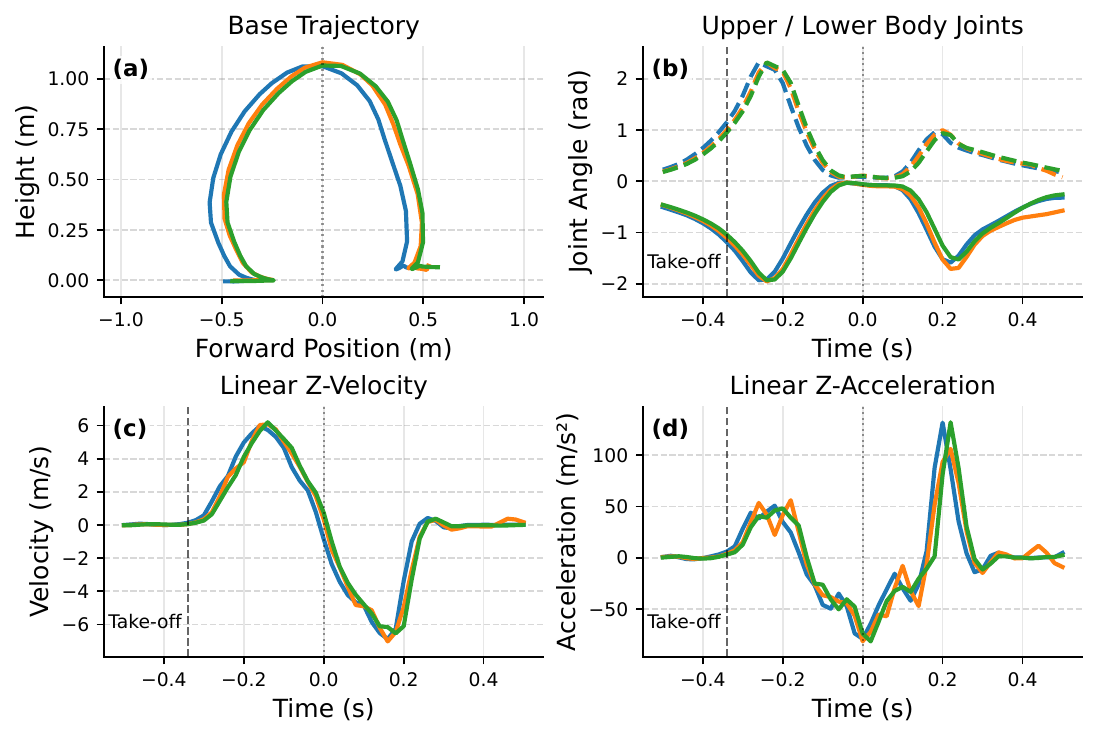}
    \caption{Three hardware flip experiments of \gls{umv}.
    The time is aligned at the peak height~($t=0$) and the interval is~$\pm 0.5$~s around that peak.
    (a) \base X-Z position, 
    (b) \lowerr and \upper joint angles, 
    (c) \base vertical velocity, and
    (d) \base vertical acceleration. 
    The vertical dashed line indicates the time at take-off. 
    }
    \label{fig:kinematics}
    \end{figure}
    
\subsection{Hardware Experiments}\label{results_sec_2}
We deployed policy~$\pi_2$ on two different~\gls{umv}s. 
Figure~\ref{fig:flip_overlay} shows the hardware experiment from one robot, 
and~\fref{fig:kinematics} shows the outcome of three different runs on another.

The robot successfully performed multiple front flips on multiple different hardware, demonstrating the policy's robustness and the effectiveness of our methodology. 
From ~\fref{fig:kinematics}(c,d), we observe that this agile maneuver is completed within~\unit[0.8]{s} of flight time, achieving a full 360-degree rotation while providing sufficient time to prepare for a controlled smooth landing with low vertical velocity and acceleration.
From \fref{fig:kinematics}(b), 
we observe that the robot extends its joints before take-off, 
tucks its boing to gain angular momentum and untucks again for smooth landing, 
and then transitions to the default joint positions,
all without reaching any joint limits. 

\subsection{Ablation: Iterative vs. Single-Shot Imitation}\label{results_sec_3}
Here, we compare two settings: \gls{imi}, which iteratively refines motion imitation sequences, against a single-iteration motion imitation baseline. 
To do so, we train two different policies with identical settings except that the first policy is trained with the initial reference~$\xi_0$ 
while the second policy is trained with the reference from the previous~\gls{imi} iteration~$\xi_1$.  
In other words, we compare training from a refined reference trajectory~$\xi_1$ against a baseline trained using an initial reference trajectory $\xi_0$.  

Figure~\ref{fig:ablation_wandb} summarizes the comparison between the two settings, averaged over three runs with different seeds. \fref{fig:ablation_wandb}(a) shows the success rate, where success is defined as episodes that end in time-outs. \fref{fig:ablation_wandb}(b,c) reports the number of environments (out of 4096) terminated at each step due to body joint limit and motor torque violations, respectively. \fref{fig:ablation_wandb}(d) shows terminations caused by touchdown velocity limit. 
%
%
Sudden drops in the success rate are moments when the termination curriculum was enforced.

Based on~\fref{fig:ablation_wandb}, with~\gls{imi}, the training converges substantially faster compared to the training from the initial reference.
Furthermore, \gls{imi} achieved a success rate of~\unit[93]{\%}, whereas training from the initial reference reaches only~\unit[49]{\%}. 
The most prominent cause of failure was due to joint limits~(\fref{fig:ablation_wandb}(b)). 
For a successful flip maneuver, the robot needs to tuck its boing as close to its limit as possible to achieve maximum angular velocity. Then, it needs to quickly untuck to prepare for a smooth landing. Doing this complicated maneuver in a limited flight time is prone to joint position and motor torque violations without rich reference guidance. The refined trajectory $\xi_1$ already anticipates the need to open its body in preparation for a safe landing compared to $\xi_0$, where smooth landing was not taken into consideration. Another prominent cause of failure was due to touch down velocity (\fref{fig:ablation_wandb}(c)), which occurs due to limited time for extending posture for smooth landing.

\subsection{Task Adaptation: Flip Up a Box}
\begin{figure}[t]
    \centering
    \includegraphics[width=0.9\columnwidth]{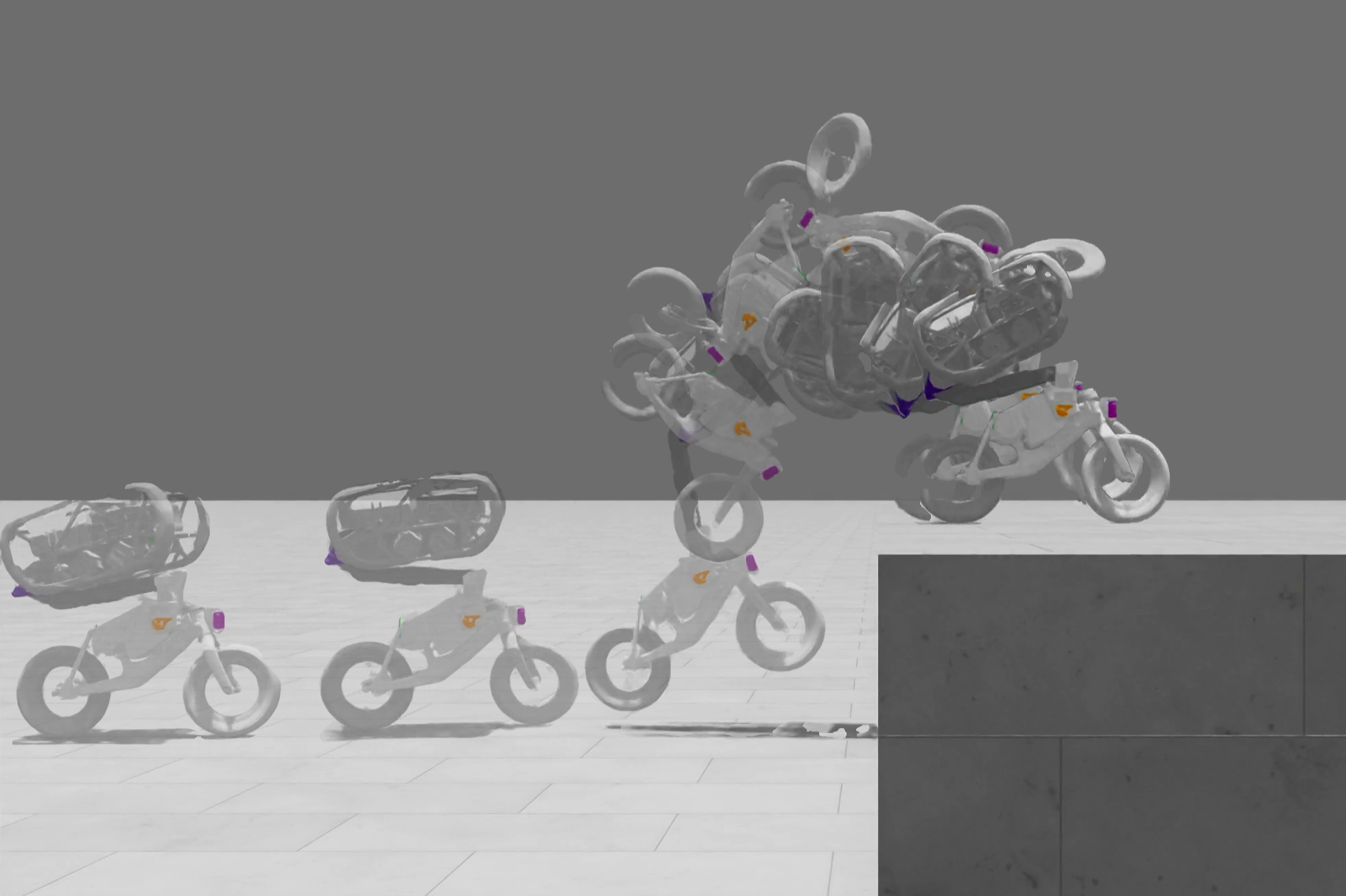}
    \caption{Overlay of \gls{umv} flipping up a \unit[70]{cm} box.}
    \label{fig:flip_up}
\end{figure}

\begin{figure}[t]
    \centering
    \includegraphics[width=1.0\columnwidth]{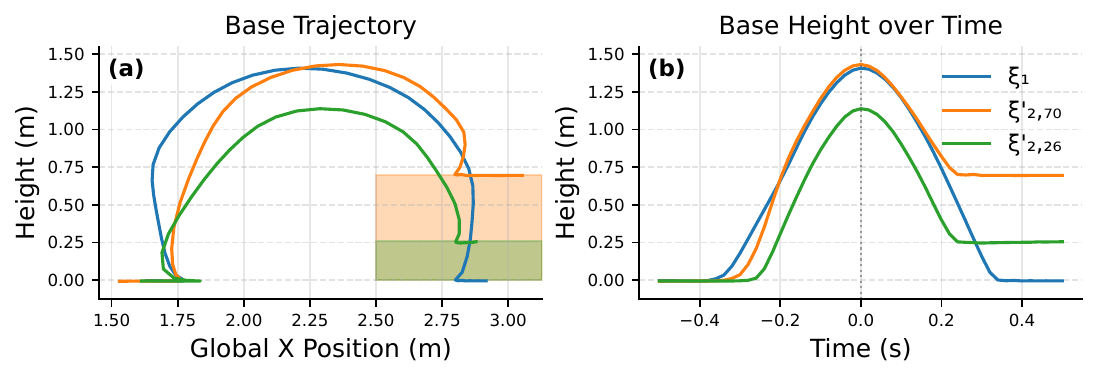}
    \caption{Flipping up \unit[70]{cm} ($\xi'_{2,70}$) and \unit[26]{cm} ($\xi'_{2,26}$) boxes, each learned by imitating the reference trajectory $\xi_1$. (a) \base trajectory in the global frame, with the shaded region indicating the table position. (b) \base vertical trajectory aligned at peak height.}
    \label{fig:flip_up_trajectory}

\end{figure}

To showcase~\gls{imi}'s effectiveness beyond safety-focused refinement, we task the robot to do a harder maneuver of flipping up a box. 
Starting from the refined reference of $\xi_1$, 
we train a separate policy~$\pi_2'$ with a modified terrain.
This training includes a~\unit[70]{cm} (as shown in~\fref{fig:flip_up}) or a~\unit[26]{cm} box in front of the robot.
As shown in~\fref{fig:flip_up}, the robot was able to successfully perform a flip-up stunt over a~\unit[70]{cm} box using~\gls{imi}.
Figure~\ref{fig:flip_up_trajectory} (a,b) shows how the robot deviates from its reference depending on the task difficulty.

\begin{table}[t]
\centering
\caption{Flip-up performance from $\xi_0$, $\xi_1$, and $\xi_2$.}
\label{tab:flipup}
\renewcommand{\arraystretch}{1.1}
\begin{tabular}{l|cc|cc}
\hline
~& \multicolumn{2}{c}{Success Rate} & \multicolumn{2}{c}{Average Return} \\
$\xi$ & 70cm & 26cm & 70cm & 26cm \\
\hline
$\xi_2$ & \textbf{95.5\%} & \textbf{97.2\%} & \textbf{81.8} & \textbf{94.1}\\
$\xi_1$ &          91.9\% &          92.4\% &              80.5 &          93.0\\
$\xi_0$ &           0.0\% &          85.1\% &          33.6 &          57.0\\
\hline
\end{tabular}
\end{table}

For quantitative analysis, we perform a third \gls{imi} iteration, where the policy is trained to execute a flip-up by imitating a trajectory generated from the second-iteration of a ground-to-ground flip $\xi_2$. As shown in the first row of \tref{tab:flipup}, this three-iteration process enhances flip-up performance in terms of both success rate measured from the start and average return. Furthermore, the \unit[70]{cm} flip-up stunt could not be achieved by directly imitating the initial reference $\xi_0$ (\ie non-iterative motion imitation), which lacks sufficient angular momentum and height clearance.

\section{Limitations and Future Work}
While successfully demonstrating robust flipping maneuvers, 
we acknowledge several limitations of~\gls{imi}.
The iterative refinement process relies on human judgment to decide when to initiate additional iterations. 
The operator qualitatively assesses the simulated trajectory's performance and feasibility, and if the outcome is unsatisfactory, a new iteration is triggered. 
%
Furthermore, the learned policy is specialized to a single skill. 
While robust within that skill, 
the resulting controller cannot perform other acrobatic maneuvers without training a separate policy, 
requiring the re-iteration of the process to guide to a different direction.

These limitations suggest promising directions for future work. 
To reduce reliance on human assessment, 
one could design automated stopping criteria that detect when the performance improvements plateau (e.g., landing stability or energy efficiency).
Moreover, an ambitious extension would be to replace the discrete iteration loop with a continuous refinement process, 
potentially leveraging~\gls{rlhf}~\cite{christiano2017deep} to incorporate the operator's preferences in a more structured and scalable manner.
To generalize beyond a single skill, we plan to extend~\gls{imi} 
to learn command-conditioned policies that learn from a library of refined skills, 
as well as investigate ways to automatically generate initial reference trajectories from high-level task objectives.

\section{Conclusion}

In this paper, we introduced~\acrfull{imi}, a reinforcement learning framework that learns agile and physically robust robotic behaviors by iteratively refining an imperfect initial reference. We demonstrated that by iteratively imitating trajectories generated by its own predecessor policies,~\gls{imi} effectively transforms a dynamically infeasible motion into a high-performance policy that respects hardware limits. Our key insight is that this recursive process naturally amplifies the effectiveness of standard motion imitation techniques such as~\acrfull{rsi} and reward shaping, enabling robust learning without complex reward engineering.

We validated our approach on the \gls{umv}, a \bicycle robot. The~\gls{imi}-trained policy successfully executed unassisted front flips on hardware, marking the first demonstration of such acrobatic stunts on this platform.

\section{Acknowledgement}
We thank Ardalan Tajbakhsh for providing the initial trajectories from the model-based controller
and the UMV team for their hardware support.

\scriptsize
\bibliographystyle{IEEEtran} 
\bibliography{./includes/bibliography.bib}
\end{document}